\documentclass{ijcaArticle}
\usepackage{amsmath}
\usepackage{amssymb}
\usepackage{indentfirst}
\usepackage{multirow}
\usepackage{booktabs}
\usepackage{svg}
\usepackage{graphicx}
\usepackage{stfloats}
\usepackage{cite}
\usepackage{subcaption}
\usepackage{authblk}
\usepackage[misc]{ifsym}
\usepackage{hyperref}
\hypersetup{
    colorlinks=true,
    linkcolor=blue,
    filecolor=blue,      
    urlcolor=blue,
    citecolor=cyan,
}

\begin{document}
\title{Cross-platform Product Matching Based on Entity Alignment of Knowledge Graph with RAEA Model} 

\author[1,2]{Wenlong Liu}
\author[2]{Jiahua Pan}
\author[2]{Xingyu Zhang}
\author[1,2]{Xinxin Gong}
\author[2]{Yang Ye}
\author[2]{Xujin Zhao}
\author[3]{Xin Wang}
\author[1]{Kent Wu}
\author[1]{Hua Xiang}
\author[2,1]{Houmin Yan}
\author[2,1,*]{Qingpeng Zhang}

\affil[1]{The Laboratory for AI-Powered Financial Technologies, Hong Kong SAR, China}
\affil[2]{School of Data Science, City University of Hong Kong, Hong Kong SAR, China}
\affil[3]{College of Intelligence and Computing, Tianjin University, China}
\affil[*]{Corresponding author: \href{mailto:qingpeng.zhang@cityu.edu.hk}{qingpeng.zhang@cityu.edu.hk}}
\renewcommand*{\Affilfont}{\small\it} %
\renewcommand\Authands{ and } %
\keywords{Product matching, Entity alignment, Knowledge graph, Graph neural network}
\maketitle 


\begin{abstract}
Product matching aims to identify identical or similar products
sold on different platforms.
By building knowledge graphs (KGs), the product matching problem
can be converted to the Entity Alignment (EA) task, which aims
to discover the equivalent entities from diverse KGs. The existing
EA methods inadequately utilize both attribute triples and relations
triples simultaneously, especially the interactions between
them. This paper introduces a two-stage pipeline consisting of rough filter and fine filter to match products from eBay and Amazon. For fine filtering, a new framework for Entity Alignment, \textbf{R}elation-aware and \textbf{A}ttribute-aware Graph Attention
Networks for \textbf{E}ntity \textbf{A}lignment (RAEA), is employed. RAEA focuses on the interactions between attributes triples and relation triples, where the entity representation aggregates
the alignment signals from attributes and relations with Attribute-aware Entity Encoder and Relation-aware Graph Attention Networks. The experimental results indicate that the RAEA model achieves significant improvements over 12 baselines on EA task in the cross-lingual dataset DBP15K (6.59\% on average Hits@1) and delivers competitive results in the monolingual dataset DWY100K. The source code for experiments on DBP15K and DWY100K is available at github\footnote{\url{https://github.com/Mockingjay-liu/RAEA-model-for-Entity-Alignment}}. 
\end{abstract}
                    
\section{Introduction}
With the development of the Internet and mobile technologies, online shopping becomes more and more popular, which leads to the prosperity of e-commerce platforms such as Amazon, eBay and JD. As different consumers prefer different platforms, it is common for sellers to sell their products on multiple online platforms simultaneously. For consumers, purchasing a product from a certain platform is usually decided based on the comparison of prices, qualities, and users’ reviews. Meanwhile, for both retailers and platform managers, comparing their products with competitors can help them to grasp market information, which helps make well-founded adjustments to pricing and sales strategy. Therefore, identifying the same products originating from different e-commerce platforms, called product matching task\cite{peeters2020intermediate}, is one of the key challenges in e-commerce applications such as price comparison portals, online marketplaces\cite{peeters2022cross}, and the building of product knowledge graphs\cite{xu2020product} such as the one constructed by Amazon\cite{dong2020autoknow}.

E-commerce platforms build knowledge graphs with their own massive accumulated data. As each e-commerce platform has different category systems, product description rules, languages, etc., the product matching can be converted to entity alignment (EA) task among cross-language heterogeneous e-commerce knowledge graphs, also called entity matching\cite{fu2020iscas_icip}.

The specific technology has developed from classical models based on KG embeddings such as JE\cite{Hao2016AJE} and MTransE\cite{Chen2017MultilingualKG} to latest models based on graph neural networks such as ClusterEA\cite{Gao2022Clusterea}. And the recent research topics of EA mainly focus on large-scale KGs\cite{Ge_2021, LIME_2022, Xin_2022, Liu_2022}, time-aware EA\cite{TEAGNN, TREA_2022}, and multi-modal EA\cite{MSNEA_2022, MCLEA_2022}.
Despite considerable explorations into the utilization of graph structure, attribute triples and relation triples, there are still challenges that only a few studies work on. First, one of the most important problems is the lack of attention to the interactions between entities and relations. Second, since the data used for building KGs is derived from different platforms, information about the same product may vary a lot in the meanings and form of description. Third, the text semantic information and graph structure information are both important in product matching, thus we need to integrate the structured as well as the unstructured information.

To achieve better results on the EA task, we propose a new GNN-based model called RAEA, which combines the advantage of AttrGNN\cite{liu2020exploring} that aggregates alignment signals from attribute values, and the advantage of RAGA\cite{zhu2021raga} that catches the mutual effect between entities and relations. We employ a two-stage pipeline to match products across the data from eBay and Amazon\footnote{\url{https://nijianmo.github.io/amazon/index.html}}, 
where the RAEA model is utilized as the fine filter. To verify the effectiveness of RAEA, we compare its performance with baselines on both DBP15K\cite{sun2017cross} and DWY100K\cite{sun2018bootstrapping} datasets, where the proposed RAEA achieves competitive performance compared with state-of-the-art methods. Applying the RAEA model in the two-stage pipeline, we achieve excellent product matching results on the eBay-Amazon dataset with NDCG = 0.5778.

An early shorter version of this paper was accepted to be presented in KGMA2022\footnote{\url{https://kgma.github.io/}}. 
As an extension paper, this paper explains more technical details, and verifies the strength of RAEA via comparison with baselines on the same methodology topic on public dataset. Additionally, we conduct ablation studies in this paper to further obtain deep insights into the contribution of different components of RAEA.

\section{Related Work}
The approach of embedding-based EA typically consists of three components, an embedding module, an alignment module, and an inference module\cite{zhang2021comprehensive}. 
The embedding module aims to learn low-dimensional vector representations. The input features may be KG structure (in the form of relation triples in the raw KG data), relation predicates and attributes. 
The alignment module aims to unify the embeddings of the two KGs into the same vector space so that aligned entities can be identified.
The inference module aims to predict whether a pair of entities from G1 and G2 are aligned.

EA approaches can be divided into two categories: translation-based approaches and GNN-based approaches. 
\begin{figure*}[h]
  \centering
  \includegraphics[width=1.0\linewidth]{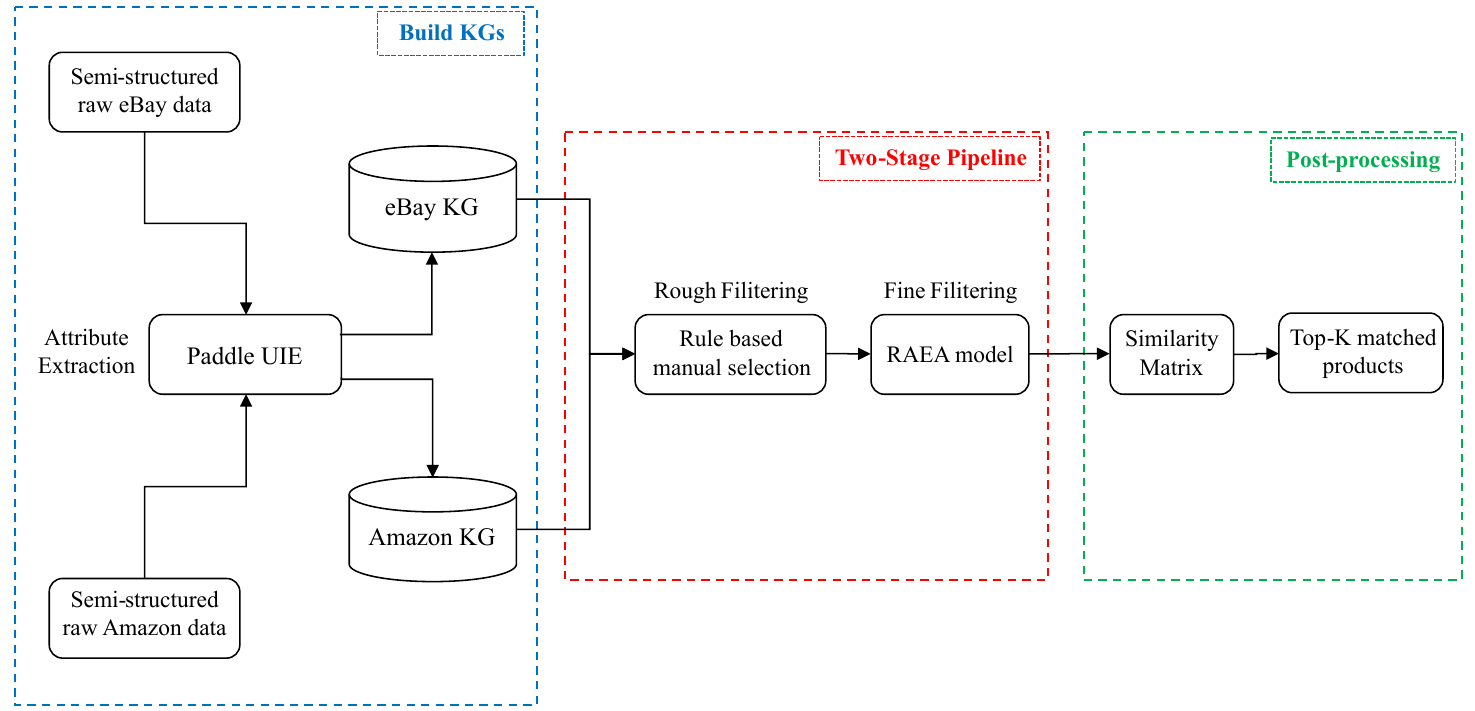}
  \caption{Flow chart for product matching cross Amazon and eBay.}
  \label{Flowchart}
\end{figure*}

\subsection{Translation-based EA Techniques}
MTransE\cite{Chen2017MultilingualKG} is the first translation-based EA model that uses TransE\cite{NIPS2013_1cecc7a7} to embed the entities and relation predicates from each KG. 
IPTransE\cite{zhu2017iterative} firstly learns the embedding of KGs separately with an extension of TransE\cite{NIPS2013_1cecc7a7} named PTransE\cite{Lin2015ModelingRP}, which can model indirectly connected entities by considering the path between them. 
BootEA\cite{sun2018bootstrapping} models the EA task as a one-to-one classification problem and the counterpart of an entity is regarded as the label of the entity. It iteratively learns the classifier via bootstrapping from both labeled data and unlabeled data. To address the deficiency of TransE\cite{NIPS2013_1cecc7a7} that its relation predicate embeddings are entity-independent, TransEdge\cite{sun2019transedge} proposes an edge-centric translational embedding model which regards the contextualized embedding of the relation predicate as the translation from the head entity to the tail entity.

The above approaches only use KG structure information, while there are some approaches that exploit relation predicates and attributes. 
JAPE\cite{sun2017cross} makes use of attribute triples and its embedding module consists of structure embedding and attribute embedding, albeit limited to only data types of the attribute values.
KDCoE\cite{Chen2018CotrainingEO} builds on top of MTransE by shifting the entity embeddings by the embeddings of entity descriptions, which are treated as a type of special attribute triples where the attribute value is a literal description for the entity.
AttrE\cite{trisedya2019entity} is the first unsupervised technique that makes use of attribute values and structure embedding. Moreover,
SDEA\cite{SDEA_2022} builds bridges between entities with attribute embedding module and relation embedding module which are driven by semantics.
\subsection{GNN-based EA Techniques}
There are growing numbers of EA techniques based on GNNs recently since GNNs suit KGs' inherent graph structure. GNN-based EA techniques usually encode KG structure by the neighborhoods of entities and many of them take attributes as input features for the embedding module because aligned entities tend to have similar neighborhoods and attributes.

GCN-Align\cite{wang2018cross} is the first study on GNN-based EA which uses vanilla GCN to model the KG structure, suffering from the structural heterogeneity of different KGs. To tackle the challenge, MuGNN\cite{Cao2019MultiChannelGN} reconciles the structural heterogeneity via multiple channels, where each channel encodes KGs with self-attention and cross-KG attention mechanism. NMN\cite{Wu2020NeighborhoodMN} captures both the neighborhood difference by estimating the entity similarities. GMNN\cite{xu2019cross} generates entity representations that contain contextual information in KG by introducing a local sub-graph of each entity. Except for the KG structure, KECG\cite{li-etal-2019-semi}, MRAEA\cite{Mao2020MRAEA} and KE-GCN\cite{Yu2021KE-GCN} utilize relation-aware GNNs to capture more neighborhood information for the EA task, but they cannot leverage multi-relation information. To address the issue, AVR-GCN\cite{ye2019vectorized} propose a vectorized relational GCN to learn the representations of entities and relations simultaneously for multi-relational networks. Besides, RePS\cite{RePS_2022} use a graph relation network that involves relation importance during aggregation to capture relations and neighborhood structure. Moreover, by incorporating the relational information into GCN with knowledge distillation, RKDEA\cite{RKDEA_2022} learns the entity representations with graph structure. To highlight the beneficial distant neighbors and reduces noises, AliNet\cite{sun2020knowledge} employs an attention mechanism and uses a gating mechanism to control the aggregation of both distant and direct neighborhood information. In addition to the KG structure and neighbor relations, some models consider leveraging the attributes of entities. For instance, AttrE\cite{trisedya2019entity} generates attribute character embeddings with attribute values. AttrGNN\cite{liu2020exploring} uses graph partition and attributed value encoder to deal with various types of attribute triples. MultiKE\cite{zhang2019multi} calculates the entity embeddings based on the perspectives from names, relations, and attributes of entities. However, most GNN-based methods ignore the interactions between attribute triples and relations triples, which cannot be fully utilized. In this work, our RAEA model leverages the attributes and relations simultaneously and explores the interaction between them for better entity embeddings.

\begin{table*}
    \centering
    \begin{tabular}{ccc}
    \toprule
    Attributes & Description & Sample\\
    \hline
    name&name of the product&Gravity Hook (Style 5)\\
    supplier&supplier of the product&Taizhou Ledao Outdoor Products Co., Ltd.\\
    store& store of the product&grape-au\\
    SKU-ID& ID of the product on eBay&TS00370-2\\
    order-ID &ID of the order on eBay&23219465\\
    partner&partner of the product &TOMSON Wuhan\\
    currency&currency type of the order&USD\\
    price &price of the product &\$9.79\\
    category 1&first level category of the product&Outdoor fitness\\
    category 2&second level category of the product &Outdoor sport\\
    category&sub-category od the product&Gravity Hook\\
    \bottomrule
    \end{tabular}
    \caption{Description and sample of eBay data.}
    \label{eBay_Data}
\end{table*}
\begin{table*}
    \centering
    \begin{tabular}{ccc}
    \toprule
    Attributes & Description & Sample\\
    \hline
    asin & ID of the product on Amazon & B011B1IHQK\\
    title & name of the product & Kirkland SignatureTM Men's Ribbed Cotton Polo-White\\
    brand&brand of the product&Kirkland Signature\\
    first date&date first listed on Amazon& July 10, 2015\\
    shipping weight&shipping weight&13.6 ounces\\
    color&color&White\\
    material&material& 95\% Cotton 5\% Spandex\\
    package dimensions&package dimensions& 10.5 x 8.7 x 3.4 inches\\
    size&size& 2X-Large\\
    category 1& first level category of the product&Sports \& Fitness\\
    category 2& second level category of the product &Sports \& Fitness\\
    category 3& third level category of the product &Golf\\
    price &price of the product &\$9.79\\
    \bottomrule
    \end{tabular}
    \caption{Description and sample of Amazon data.}
    \label{Amazon_Data}
\end{table*}

\section{Methodology}
Our task is to align Amazon products with eBay products. If we regard this task as a searching or matching problem, we take the product information of platform A as the query condition, search for the matched products in platform B, and sort the candidate results of the query. Specifically, since we can obtain much more data on Amazon products than eBay products, we take eBay products as the query condition and Amazon products as the candidate set of the query results. As the flowchart is shown in Fig\ref{Flowchart}, all work is divided into three parts, consisting of building KGs, a two-stage pipeline and post-processing. Firstly, we extract product attributes from semi-structured data with Paddle UIE\footnote{\url{https://github.com/PaddlePaddle/PaddleNLP/tree/develop/model_zoo/uie}}, which are combined with the structured data to build the KGs. The two-stage pipeline for product matching includes rough filtering and fine filtering. In the rough filtering stage, the categories of products and keywords in the product title are matched based on rules, from where we get the Amazon product candidate set corresponding to each eBay product. Then, we refine the matching results with the GNN-based RAEA model in the fine filtering stage, which outputs the product similarity matrix. Finally, the Top-K best matched Amazon products for each eBay product are output after the post-processing. In this section, our method for product matching is described in detail.

\subsection{Dataset}

\subsubsection{Data Description}
The dataset contains 957,216 items of product information from the Amazon platform\cite{ni2019justifying} and 46,066 items of product information from the eBay platform. The Amazon data contains product descriptions, prices, sales rank, brand info, and co-purchasing links. The eBay data is the order records, which contain store name, partner, supplier, and description of the order. The description and samples of eBay data and Amazon data are displayed in Table \ref{eBay_Data} and Table \ref{Amazon_Data}.
\subsubsection{Data Processing}
From structured data, we extract product titles and categories. Other attributes such as product colors, size, and target customer are extracted from unstructured text data based on regular expression rules. It is worth noting that all the transaction data from eBay is in Chinese. Following Sun et al.\cite{liu2020exploring}, we use Google Translate to translate Chinese data of eBay product to English, which makes it possible to embed information about eBay products and Amazon products into the same vector space. According to the data described in Table \ref{eBay_Data} and Table \ref{Amazon_Data}, we build the KGs for eBay and Amazon respectively by constructing relation triples and attribute triples, which are used in the fine filtering stage.
\begin{figure*}[h]
  \centering
  \includegraphics[width=1.0\linewidth]{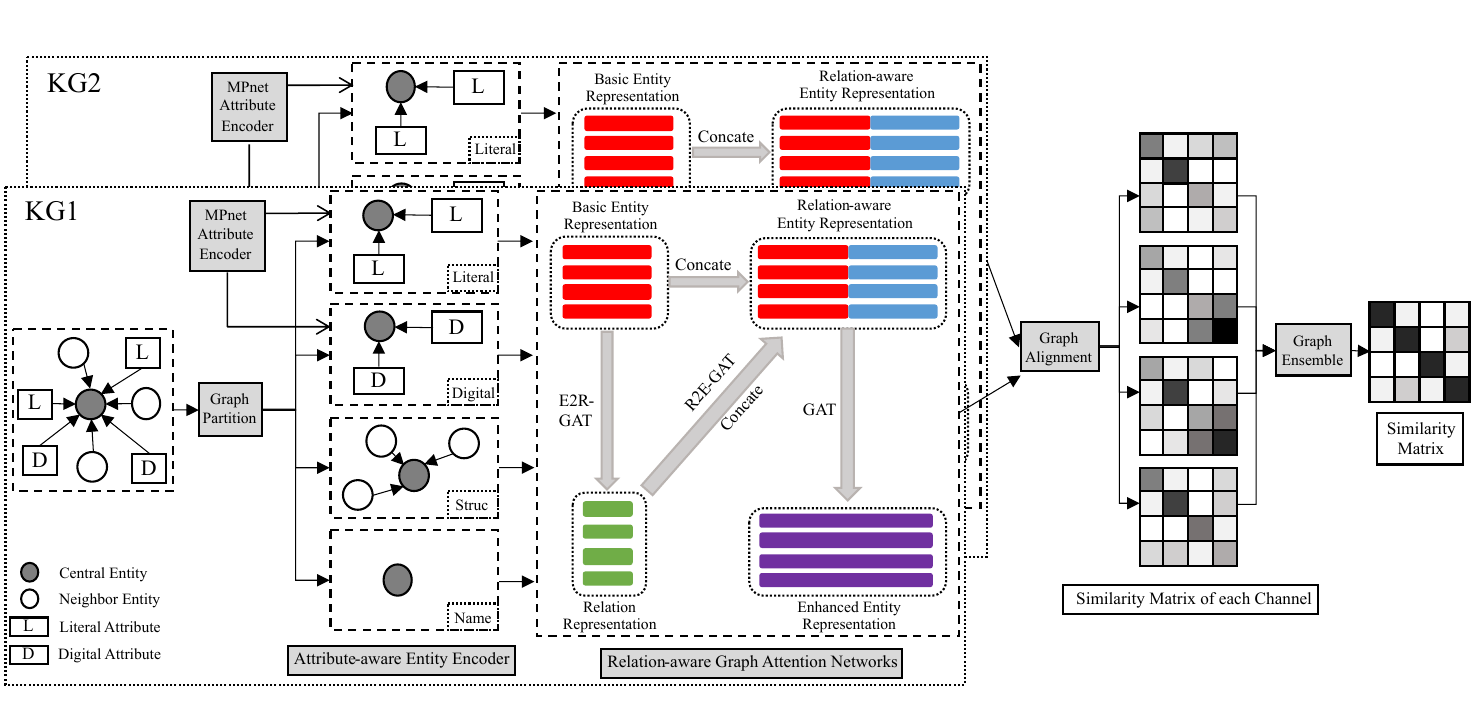}
  \caption{The framework of RAEA model.}
  \label{Framework}
\end{figure*}

\subsection{Rough Filtering}
\subsubsection{Rule-based Rough Filtering}
To narrow down the number of amazon candidates matched to each eBay product, we use a rough filter based on matching rules. Specifically, based on product names and categories, we use regular expressions to match products across two platforms. According to the observation and statistics, the granularity of the stratification is not the same on both platforms, which is hard to align categories directly. Thus, we roughly match eBay and Amazon products with the analogical categories by regular expressions.
Specifically, we simulate the process of a user searching for a specific eBay product in the Amazon database. The keywords in the matching criteria need to be able to identify the subject of the product. Therefore, we concatenate hierarchical categories for each product. Besides, Amazon product title is also concatenated after categories as additional information. The matching rules are designed manually based on expert knowledge. 
\subsubsection{Case Study of Rough Filtering}
The keywords within the regular expressions are from the categories and titles of the Amazon product. For example, for eBay products in the concatenated category "Outdoor fitness, mountaineering, rock climbing, ice climbing equipment, anti-skating claw", the Amazon products that satisfy the regular expression“climbing.*crampons” is matched. Partial roughly matched Amazon products are shown in Table \ref{rough_filter_case}. The total number of candidate Amazon products is 1,171. After rough filtering, the candidate Amazon products for each eBay product is smaller in amount.
\begin{table*}
    \centering
    \begin{tabular}{ccc}
    \toprule
    Roughly Matched Amazon Products\\
    \hline
    GRIVEL air tech light new-matic crampons\\
    black anti-slip pair ice snow shoe spikes grips crampons hiking fishing climbing\\
    STUBAI ultralight crampons pro\\
    ESUMIC anti-slip shoes ice gripper cleats crampons with pouch carabiner\\
    ice traction universal slip-on stretch fit snow ice spikes (grips, crampons, cleats) size L (black)\\
    KAHTOOLA steel hiking crampons\\
    \bottomrule
    \end{tabular}
    \caption{Partial roughly matched Amazon products for eBay products in the categories "Outdoor fitness, mountaineering, rock climbing, ice climbing equipment, anti-skating claw".}
    \label{rough_filter_case}
\end{table*}
\subsection{Fine Filtering}
\subsubsection{Model Introduction}
For Amazon candidates of each eBay product, we try to select the best matched Top-K products with the proposed EA model, RAEA. For better results of the EA task, the RAEA model is designed by introducing relation-aware graph attention networks derived from RAGA into the AttrGNN model, which captures interactions of entities and relations, and remains the diversity of relation types between two entities. The Bert used in AttrGNN is replaced by MPnet\cite{NEURIPS2020_c3a690be} based encoder, which is trained with the unsupervised SimCSE method. Besides, RAEA uses the proposed strategy to ensemble the outputs of channels, called pre-weighted, which makes important channel plays a more crucial role in the final similarity matrix of entities.
\subsubsection{Components of RAEA}
As depicted in the Fig\ref{Framework}, the framework of RAEA model contains six main functional components, including graph partition, MPnet attribute encoder, attribute-aware entity encoder, relation-aware graph attention networks, graph alignment and graph ensemble, which are explained in detail as follows.
\textbf{\\Graph Partition}.
Inspired by AttrGNN\cite{liu2020exploring}, we divided the graph into four subgraphs based on structure and types of the attribute value, where four channels are designed according to different perspectives for aggregating alignment signals. The following defines the four subgraphs:
\begin{enumerate}
\item[$\bullet$]$\boldsymbol{G}^{Lite}$ only includes attribute triples of literal values. 
\item[$\bullet$]$\boldsymbol{G}^{Digi}$ only includes attribute triples of digital values.  
\item[$\bullet$]$\boldsymbol{G}^{Name}$ only includes attribute triples of the names.
\item[$\bullet$]$\boldsymbol{G}^{Stru}$ has no attribute triples.
\end{enumerate}
The attribute triples of all the subgraphs are mutually exclusive, although they share identical relation triples.
\textbf{\\MPnet Attribute Encoder}.
We utilize MPnet to encode attributes, which combines the advantages of BERT\cite{Devlin2019BERTPO} and XLNet\cite{NEURIPS2019_dc6a7e65}. The pre-training objective function for BERT and XLNET are MLM and PLM respectively, MLM can get the position information of the whole sentence, but cannot model the dependency between prediction tokens, while PLM can model dependencies between prediction tokens through autoregressive modeling, it cannot see the position information of the whole sentence. MPnet incorporates their pre-training objective function and is able to model the dependencies between prediction tokens. Moreover, in order to allow our entities to aggregate more informative attribute values, we want our encoder to have good enough semantic similarity performance. Some studies such as BERT-flow\cite{Li2020OnTS} have found that the word representations of BERT\cite{Devlin2019BERTPO} can be constrained by the learned anisotropic word embedding space, resulting in poor performance on the semantic similarity task. To address this issue, we pre-train the MPnet with contrast learning. Specifically, we use unsupervised SimCSE \cite{gao2021simcse}. It feeds the same sentence twice and uses different hidden dropout masks to get two similar but not the same embeddings as positive pairs and the rest as negative pairs for each text entered, from where we get better semantic similarity performance.
\textbf{\\Attribute-aware Entity Encoder}.
To aggregate the attribute triples of entities, for the literal and digital channels, we use the pre-trained MPnet to obtain the features of both literal and digital values, and use the attention mechanism from GAT \cite{Chen2021rGATRG} to selectively aggregate discriminative information from attributes and values to a single vector representation of the central entity. If the initial vector representation of entity $e$ is ${h_e^0}$, the $i_{th}$ layer hidden state of entity $e$ can be defined as $\boldsymbol{h}_e^i$ as follows:
\begin{equation}
\begin{aligned}
\boldsymbol{h}_e^i = \sigma\left(\sum_{j=1}^n\alpha_{j}W_{i}\left[\alpha_{j};v_{j}\right]\right),\\
\alpha_{j} = softmax(o_j) = \frac{\exp{\left(o_j\right)}}{\sum_{k=1}^{n}\exp{\left(o_k\right)}},\\
o_j = LeakyReLU(\boldsymbol{u}^T\left[\boldsymbol{h}_{e}^{i-1};a_{j}\right])
\end{aligned}
\end{equation}
where $j\in\{1,…,n\}$, $W_{i}\in\mathbb{R}^{D_{h_i}\times(D_a+D_v)}$ and ${u\in\mathbb{R}^{(D_e+D_a)\times1}}$ are learnable matrices, $\sigma$ is the $ELU(\cdot)$ function.
\textbf{\\Relation-aware Graph Attention Networks}.
With the Attribute-aware entity encoder, we have obtained the attribute-aware entity representation with no informative relation feature. Furthermore, we hope to pass the entity presentation through relations so that the entity representation will be more accurate. Inspired by the RAGA\cite{zhu2021raga}, we apply three steps of entity to relation, relation to entity and entity enhancement to aggregate the relation features into entity representation.
\begin{enumerate}
\item[$\bullet$]\text{\itshape{Entity to Relation Representation}}.\\
Since the number of relation triples is significantly more than the number of entities, the information of KGs are heavily hidden in relationship triples. Therefore, the relation representation is potentially contributive to more accurate entity representation, which helps improve the entity alignment. We use attention mechanism to discriminatively gather information from different head entities and tail entities of certain relation. For each relation, we generate relation representation from the view of head entities and view of tail entities respectively, which will be concatenated to obtain relation representation. Different from AttrGNN, which simply regards the various relations of two entities as a single connection of two entities, the Relation-aware Graph Attention Networks calculates embedding for each relation, no matter what entities are connected. Therefore, the learned relation representations reserve the variety of relations.
For relation $r_k$, the head-view relation embedding is computed as follows:
\begin{equation}
\begin{aligned}
\boldsymbol{\alpha}_{ijk} = 
\frac{exp\left(LeakReLU\left(\boldsymbol{a}^{T}{\left[\boldsymbol{x}_{i}\boldsymbol{W}^{h}||\boldsymbol{x}_{j}\boldsymbol{W}^{t}\right]}\right)\right)}
{\begin{tiny}
{\sum_{{e_{i^\prime}}\in{H}_{r_k}}}{\sum_{{e_{j^\prime}}\in{T}_{{e_i}{r_k}}}}
exp\left(LeakReLU\left(\boldsymbol{a}^{T}{\left[{{\boldsymbol{x}}_{i^\prime}}\boldsymbol{W}^{h}||{\boldsymbol{x}_{j^\prime}}\boldsymbol{W}^{t}\right]}\right)\right)
\end{tiny}}
\end{aligned}
\end{equation}
\begin{equation}
    \boldsymbol{r}_k^h = ReLU\left({\sum_{{e_i}\in{H}_{r_k}}}{\sum_{{e_j}\in{T}_{{e_i}{r_k}}}}{\boldsymbol{\alpha}_{ijk}}{\boldsymbol{x}_i}{\boldsymbol{W}^h}\right)
\end{equation}
where $\alpha_{ijk}$ is the attention weight from head entities $e_i$ to relation $r_k$. $H_{r_k}$ is the set of head entities and $T_{{e_i}{r_k}}$ is the set of tail entities for head entity $e_i$ and relation $r_k$. Similar, we obtain the tail-view relation embedding $r_k^t$, and concatenate it with $r_k^h$ to obtain the relation representation $r_k$:
\begin{equation}
r_k= r_k^h+ r_k^t
\end{equation}
\item[$\bullet$]\text{\itshape{Relation-aware Entity Representation}}.\\
Based on the realization that the entity representation that contains information from its neighbor relations is more accurate that only contains itself, we try to gather the embedding of relations to the central entity embedding. Similar to “entity to relation representation”, we can also leverage neighbor relation embeddings to update the entity representation. 
For entity $e_i$, the out-relation embedding $x_i^h$ is computed as follows:
\begin{equation}
\begin{aligned}
\boldsymbol{\alpha}_{ik} = 
\frac{exp\left(LeakReLU\left(\boldsymbol{a}^{T}{\left[\boldsymbol{x}_{i}||\boldsymbol{r}^{k}\right]}\right)\right)}
{\begin{tiny}
    {\sum_{{e_{j}}\in{T}_{e_i}}}
    {\sum_{{r_{k^\prime}}\in{R}_{{e_i}{e_j}}}}
    exp\left(LeakReLU\left(\boldsymbol{a}^{T}{\left[\boldsymbol{x}_{i} ||\boldsymbol{r}_{k^\prime}\right]}\right)\right)
\end{tiny}}
\end{aligned}
\end{equation}
\begin{equation}
    \boldsymbol{x}_i^h = ReLU\left({\sum_{{e_i}\in{T}_{e_i}}}{\sum_{{r_k}\in{R}_{{e_i}{e_j}}}}{\boldsymbol{\alpha}_{ik}}{\boldsymbol{r}_k}\right)
\end{equation}
where $\alpha_{ik}$ is the attention weight from relation $r_k$ to head entities $e_i$. $T_{e_i}$ is the set of tail entities for head entity $e_i$ and $R_{{e_i}{e_j}}$ is the set of relations between head entity $e_i$ and relation $e_j$.
Similar, we obtain the in-relation embedding $x_i^t$, and concatenate it with $x_i^h$ and $x_i$ to obtain the relation-aware entity representation $x_i^{rel}$:
\begin{equation}
    x_i^{rel} = [x_i ||x_i^h||x_i^t]
\end{equation}
\item[$\bullet$]\text{\itshape{Entity Enhancement}}.\\
In relation-aware entity representation, the entities only take the information of one-hop relations into consideration. We use one layer of ordinary graph attention networks to enhance the information of two-hop relations. For entity $e_i$, the enhanced representation $x_i^{out}$ can be defined by:
\begin{equation}
\begin{aligned}
\boldsymbol{\alpha}_{ij} = 
\frac{exp\left(LeakReLU\left(\boldsymbol{a}^{T}{\left[\boldsymbol{x}_{i}^{rel}||\boldsymbol{x}_{j}^{rel}\right]}\right)\right)}
{\begin{tiny}{\sum_{{j^\prime}\in{N}_{i}}}
exp\left(LeakReLU\left(\boldsymbol{a}^{T}{\left[{\boldsymbol{x}}_{i}^{rel}||{{\boldsymbol{x}}_{j^\prime}}^{rel}\right]}\right)\right)
\end{tiny}}
\end{aligned}
\end{equation}
\begin{equation}
    \boldsymbol{x}_i^{out} = \left[\boldsymbol{x}_i^{rel}|| ReLU\left({\sum_{j\in{N}_{i}}}{\boldsymbol{\alpha}_{ij}}{\boldsymbol{x}_i^{rel}}\right)\right]
\end{equation}
where ${\alpha_{ij}}$ is the attention weight from entity $x_i^{rel}$ to its neighbor entities.
\end{enumerate}
\textbf{\\Graph Alignment Training}.
To evaluate whether two entities are aligned, we need to calculate the distance between the representations of the two entities. Thus, the two KG’s representations in each channel should be unified into the same vector space, which is our goal of training. Firstly, inspired by Li et al. \cite{li2019semi}, we generate negative samples $e_\_$ (or $e_\_^\prime$) for each pair of seed equivalent entities ($e$,$e^\prime$) by searching the nearest samples of $e$ (or $e^\prime$) in embedding space. Then, we reduce the distance between the seed equivalent entities and aggrandize the distance between negative samples. For each channel, we optimize the following objective function:
\begin{equation}
\begin{aligned}
    \boldsymbol{L}_k = {\sum_{(e,e^\prime)\in\psi^s}}\left(
    {{\sum_{{e_{\_}\in{NS\left(e\right)}}}}{\left[d\left(e^k,{e^\prime}^k\right)-d\left(e_{\_}^k,{e^\prime}^k\right)+\gamma\right]}_{+}} + \right. \\ \left. 
    {{\sum_{{e_{\_}^\prime\in{NS\left(e^\prime\right)}}}}{\left[d\left(e^k,{e^\prime}^k\right)-d\left(e,{{e^\prime}_{\_}}^k\right)+\gamma\right]}_{+}}
    \right)
\end{aligned}
\end{equation}
where $\psi^s$ is the set of seed equivalent entities, $NS\left(e\right)$ is the set of negative entities for entity e, and $NS\left(e^\prime\right)$ is the set of negative entities for entity $e^\prime$.
\textbf{\\Channel Ensemble}.
Except for the two strategies (Average Pooling and SVM) mentioned in AttrGNN\cite{liu2020exploring} can be used to ensemble the output from different channels to construct the final similarity matrix of entities, we propose the pre-weighted strategy. It calculates the importance weight of each channel before the channel ensemble, since different channel has different importance. We test the Hits@1 of entity alignment with the similarity matrix of each channel. Then, the importance of each channel is the ratio of Hits@1 over the sum for Hits@1 of four channels.
\begin{table*}[ht]
\centering
\setlength\tabcolsep{3.5pt}
\begin{tabular}{l|ccc|ccc|ccc|ccc|ccc} 
\toprule
\multirow{2}{*}{Methods}    & \multicolumn{3}{|c}{DBP15K (zh-en)} & \multicolumn{3}{|c}{DBP15K (ja-en)}  & \multicolumn{3}{|c}{DBP15K (fr-en)}& \multicolumn{3}{|c}{DWY100K (wd)}&\multicolumn{3}{|c}{DWY100K (yg)}\\& H@1         & H@10        & MRR            & H@1         & H@10        & MRR & H@1 & H@10        & MRR & H@1 & H@10        & MRR & H@1         & H@10        & MRR\\
\hline
MTransE\cite{Chen2017MultilingualKG} & 30.81 & 61.44 & 0.364 & 27.91          & 57.52           & 0.349          & 24.38          & 55.58          & 0.335  & 28.12	& 51.95	& 0.363	& 25.15	& 49.29	& 0.334 \\
JAPE\cite{sun2017cross}                        & 41.18          & 74.46          & 0.490           & 36.25          & 68.52           & 0.476          & 32.29          & 66.72          & 0.429   & 31.84	& 58.88	& 0.411	& 23.57	& 48.41	& 0.320\\
BootEA\cite{sun2018bootstrapping}  & 62.94          & 84.75          & 0.703          & 62.23          & 85.39          & 0.701          & 65.30           & 87.44          & 0.731  & 74.79	& 89.84	& 0.801	& 76.10	& 89.44	& 0.808 \\
TransEdge\cite{sun2019transedge}       &73.51 &91.92 &0.801 &71.86 &\underline{93.15} &0.795 &71.01 &94.14 &0.796 &78.80	&93.80	&0.824	&79.20	&93.60	&0.832 \\
\hline
MuGNN\cite{Cao2019MultiChannelGN} & 49.40           & 84.40           & 0.611          & 50.10           & 85.70           & 0.621          & 49.50           & 87.00             & 0.621    & 61.60	& 89.70	& 0.714	& 74.10	& 93.70	& 0.810 \\
AliNet\cite{sun2020knowledge}                    & 55.23          & 85.21             & 0.657         & 53.91          & 82.63          & 0.628          & 54.95          & 83.13          & 0.645    & 69.00	& 90.80	& 0.766	& 78.60	& 94.30	& 0.841\\
RDGCN\cite{Wu2019RelationAwareEA} & 70.75          & 84.55          & 0.749          & 76.74          & 89.54          & 0.812          & 88.64          & 95.72          & 0.908 & 97.90	& 99.10	& - 	& 94.70	& 97.30	& - \\
Dual-AMN\cite{10.1145/3442381.3449897} &73.10	&92.30	&0.799	&72.60	&92.70	&0.799	&75.60	&94.80	&0.827	&78.60	&95.20	&0.848	&86.60	&97.70	&0.907\\
MultiKE\cite{zhang2019multi}    & 43.70           & 51.62          & 0.466          & 57.00             & 64.26          & 0.596          & 71.43          & 76.08          & 0.733  & 91.86	& 96.26	& 0.935	& 82.35	& 93.30	& 0.862 \\
AttrGNN\cite{liu2020exploring} & \underline{79.60}           & \underline{92.93}          & \underline{0.845}          & 78.33          & 92.08          & 0.834          & \underline{91.85}          & \underline{97.77}          &\underline{0.910} & 96.08	& 98.86	& \underline{0.972}	& \textbf{\underline{99.89}}	& \textbf{\underline{99.99}}	& \textbf{\underline{0.999}} \\
NMN\cite{Wu2020NeighborhoodMN} &73.30	&86.90	&0.702	&\underline{78.50}	&91.20	&\underline{0.870}	&90.20	&96.70	&0.864	&\textbf{\underline{98.10}}	&\underline{99.20} &-  &96.00	&98.20	&-\\
RKDEA\cite{RKDEA_2022} &60.30	&87.20	&0.703	&59.70	&88.10	&0.698	&62.20	&91.20	&0.721	&75.60	&97.30 &0.821  &82.30	&97.10	&0.879\\
\hline
${RAEA}^{avg}$                   & 82.49          & 95.95          & 0.875          & 82.98          & 96.28          &0.879          & 85.82          & 97.49          & 0.902  &94.97	&99.09	&0.966	&99.33	&99.88	&0.996         \\
${RAEA}^{svm}$                   & 78.00             & 94.14          & 0.841          & 77.34          & 93.63          & 0.834          & 76.23          & 93.65          & 0.827     &91.14	&97.94	&0.937	&99.46	&99.90	&0.996\\
\textbf{RAEA} & \textbf{86.28} & \textbf{97.07} & \textbf{0.903} & \textbf{88.48} & \textbf{97.71} & \textbf{0.920} & \textbf{94.97} & \textbf{99.60} & \textbf{0.968}   & 97.35	 & \textbf{99.72}	 & \textbf{0.983}	 & 99.82	 & 99.98 & \textbf{0.999}\\
\hline
{\itshape{Improv. best}} & 6.68             & 4.14          & 0.058          & 9.98          & 4.56          & 0.050          & 3.12          & 1.83          & 0.058     &-0.75	&0.52	&0.011	&-0.07	&-0.01	&0\\
\bottomrule
\end{tabular}
\caption{Performance of Translation-based model, GNN-based model and the proposed RAEA on DBP15K and DWY100K datasets. The results of baselines are collected from the origin papers. ${RAEA}^{avg}$ and ${RAEA}^{svm}$ use the average pooling and SVM as ensemble strategy respectively. The RAEA uses the proposed pre-weighted strategy to ensemble the channels. The number underlined is the maximum value in baselines. The number in bold is the maximum value in each column.}
\label{Table_result}
\end{table*}

\section{Experiments}
\begin{table}
\setlength\tabcolsep{1.5pt}
    \centering
    \begin{tabular}{l|cc|cc|cc|cc|cc}
    \toprule
    \multirow{2}{*}{Methods}  & \multicolumn{8}{c}{eBay-Amazon} \\
    &  \multicolumn{2}{|c}{NDCG} & \multicolumn{2}{|c}{Recall} & \multicolumn{2}{|c}{Precision} & \multicolumn{2}{|c}{MRR} \\
    \hline
    ${RAEA}^{avg}$ & \multicolumn{2}{|c}{0.563} & \multicolumn{2}{|c}{0.856} & \multicolumn{2}{|c}{0.542} & \multicolumn{2}{|c}{0.341}\\
    95\% CI & 0.546 & 0.571 & 0.843 & 0.873 & 0.526 & 0.560 & 0.331 & \multicolumn{1}{c}{0.349}\\
    \hline
    RAEA & \multicolumn{2}{|c}{0.566} & \multicolumn{2}{|c}{0.862} & \multicolumn{2}{|c}{0.558} & \multicolumn{2}{|c}{0.345}\\
    95\% CI & 0.542 & 0.576 & 0.830 & 0.893 & 0.529 & 0.571 & 0.324 & \multicolumn{1}{c}{0.358}\\
    \bottomrule
    \end{tabular}
    \caption{Performance of RAEA on dataset eBay-Amazon.}
    \label{tab:my_label}
\end{table}
We apply the two-stage pipeline to the dataset of eBay-Amazon products to observe the effectiveness of the method in real-world data. Furthermore, to certify that the proposed RAEA model used in the fine filtering stage is a state-of-the-art method on the EA task, we evaluate the performance of RAEA for 3 different channel ensemble strategies and compare them with 10 baselines on two public datasets DBP15K\cite{sun2017cross} and DWY100K\cite{sun2018bootstrapping} for EA task, which shows the RAEA model outperforms the state-of-the-art methods.
\subsection{Experiments on eBay-Amazon Dataset}
\subsubsection{Experimental Settings}
\textbf{\itshape{\\Evaluation Metrics}}.
The task in the fine-filtering stage can be regarded as a searching or matching problem, where we take eBay products as the query condition and Amazon products as the candidate set of the query. Thus, the ranking measurement of candidates reflects the effectiveness of fine-filtering.
We employ the Normalized Discounted Cumulative Gain (NDCG) as the evaluation
matrix of the experiments, which is widely used to evaluate the ranking in search engine, recommendation system, and expert finding\cite{wang2013theoretical}. Higher NDCG means a preferable ranking of matched Amazon candidate products, indicating better performance on EA task. The NDCG has two advantages. First, NDCG evaluates the degree of relevancy with the query for each candidate in the ranking of the query results. Second, NDCG contains a discount function, taking the position of each candidate in the ranking into account, which is definitely important since we care the top-ranked query results much more than others\cite{wang2013theoretical}. Besides, we also use recall, precision and mean reciprocal rank (MRR) as the assistant evaluation matrix. To evaluate the stability of the model performance, we use bootstrapping approach and report the 95\% confidence interval in each metric and each model in Table \ref{tab:my_label}.\\
\textbf{\itshape{Training Details}}.
When training the encoder with unsupervised SimCSE, we use “sentence-transformers/paraphrase-multilingual-MPnet-base-v2”\footnote{\url{https://huggingface.co/sentence-transformers/paraphrase-multilingual-MPnet-base-v2}}\cite{reimers-2019-sentence-bert} as the pre-trained model, where the batch size is set to 128.  We choose Adagrad\cite{duchi2011adaptive} as the optimizer when training RAEA model. We generate 15 negative samples for each pair of seed equivalent entities in Graph Alignment. There are three GNN channels, which are Literal, Digital, and Structure channels. We abandon the Name channel since the name of each entity is a standard identification number, which contains no information of the product, and the name of the product is taken as a literal attribute in the Literal channel. When training each channel, we utilize grid search to find the best parameters, with the following ranges: learning rate \{1e-3, 4e-3, 7e-3\}, L2 regularization \{0, 1e-4, 1e-3\}. Each channel is trained at most 1500 epochs, but it may early stop if no higher Hits@1 exists in more than continuous 50 epochs.
\subsubsection{Results Analysis}
According to Table \ref{tab:my_label}, NDCG of both RAEA(avg) and RAEA(pre-weighted) is about 0.56, which means the candidates of matched Amazon products for eBay products are kind of acceptable. Observing the 95\% confidence interval of each metric, we find that the performance of our method is stable. To obtain higher NDCG on product matching, we need more annotated seed equivalent entities with large size and high quality because our method is supervised learning. However, since the annotated seed equivalent entities, manually annotated by non-specialists, are of small number and unreliable, the differences between the two ensemble strategies are insignificant. Therefore, it's necessary to verify the performance of the RAEA model on the public dataset used for entity alignment, expecting excellent performance on public datasets makes it convincing that RAEA plays a contributive role in the fine filtering stage.
\subsection{Experiments on Public Dataset}
\subsubsection{Experimental Settings}
\textbf{\itshape{\\Dataset}}. We test all models on a cross-lingual dataset DBP15K\cite{sun2017cross} and a monolingual dataset DWY100K\cite{sun2018bootstrapping}. DBP15K contains three cross-lingual KGs collected from DBpedia: Chinese and English, Japanese and English, French and English. DWY100K includes DBpedia and Wikidata(DBP-WD), DBpedia and YAGO(DBP-YG), which are monolingual datasets. Following Liu et al.\cite{liu2020exploring}, the attribute triples are retrieved from DBpedia dump(2016-10). For each KG, we randomly sample 30\% of equivalent entity pairs for training, the rest for testing. For DWY100K, we use the same training/testing split with \cite{zhang2019multi}, where we randomly sample 30\% of equivalent entities pairs for training, 10\% for validation, and the rest for testing. \\
\textbf{\itshape{\\Baselines}}. To evaluate the effectiveness of the proposed model, we compare the performance of RAEA with the selected 12 baseline models based on different alignment methods:
\begin{itemize}
\item[-] Translation-based: MTransE\cite{Chen2017MultilingualKG}, JAPE\cite{sun2017cross}, BootEA\cite{sun2018bootstrapping}, TransEdge\cite{sun2019transedge}
\item[-] GNN-based: MuGNN\cite{Cao2019MultiChannelGN}, AliNet\cite{sun2020knowledge},  RDGCN\cite{Wu2019RelationAwareEA}, Dual-AMN\cite{10.1145/3442381.3449897}, MultiKE\cite{zhang2019multi}, AttrGNN\cite{liu2020exploring}, NMN\cite{Wu2020NeighborhoodMN}, RKDEA\cite{RKDEA_2022}
\end{itemize}
Note that MultiKE\cite{zhang2019multi} and AttrGNN\cite{liu2020exploring} incorporate the information of entity name, attributes and relations, which are the same as the proposed RAEA model. But other competitive models do not use so much information, such as AliNet\cite{sun2020knowledge} solely relies on structure information. Note that since some recent models differ sharply from our model in data and methods used for entity alignment, such as MCLEA\cite{MCLEA_2022} based on multi-modal data and non-neural framework LightEA\cite{LightEA_2022}, these models aren't taken into comparison.\\
\textbf{\itshape{Evaluation Metrics}}.
As for evaluation metrics, we employ Hits@K(\%, K=1 and K=10 in our experiments) and Mean Reciprocal Rank(MRR) to evaluate the performance. Higher Hits@K and MRR manifest better performance on EA task. Note that, the Hits@1
is equivalent to precision, recall, and F1-measure\cite{sun2020knowledge}.
\textbf{\itshape{\\Training Details}}.
For the DBP15K dataset, most training details are the same as the experiments on the eBay-Amazon dataset, except for three differences. First, the batch size of pre-trained model\cite{reimers-2019-sentence-bert} is set to 512. Second, there are four GNN channels, including Literal, Digital, Structure, and Name channels. For each channel, the ranges of grid search are as follows: L2 regularization \{0, 1e-4, 1e-3, 1e-2\}, learning rate \{1e-3, 5e-3, 1e-2, 15e-3, 2e-2\}. With repeating grid search and narrowing the ranges of searching hyperparameters, we find the best hyperparameters for the Literal/Digital channel is (L2 regularization = 0, learning rate = 15e-3), for Structure channel is (L2 regularization = 1e-2, learning rate = 1e-2), and for Name channel is (L2 regularization = 0, learning rate = 5e-3). Since the DWY100K is much larger than the DBP15K in scale, we only generate 5 negative samples for each pair of seed equivalent entities to save time and space for training on DWY100K. We conduct our experiments on a server with one 12-core 3.50GHz CPU and one NVIDIA GeForce RTX 3090 GPU with 24GB memory. On DBP15K datasets, the Literal/Digital/Structure channel costs about 25 minutes for a grid search, and the Name channel costs less than 5 minutes. As for DWY100K datasets, the grid search in Literal/Digital/Structure channel costs less than 70 minutes, and the Name channel costs less than 10 minutes. The theoretical time complexity of training each channel can be represented as follows:
\begin{equation}
T = P_1P_2(e+\frac{e}{F_{neg}}N_{neg})N_sN_tT_sT_t
\end{equation}
where $P_1$ and $P_2$ are the sizes of the hyperparameters range of L2 regularization and learning rate respectively in a grid search. The number of epochs is represented as $e$. $F_{neg}$ is the frequency of updating the negative samples, which means we generate $N_{neg}$ negative samples again after training $F_{neg}$ epochs. $N_s$ and $T_s$ are the numbers of nodes and triples in the source graph and $N_t$ and $T_t$ are the numbers of nodes and triples in the target graph. For DBP15K datasets, there are about 0.5 million trainable parameters in the Literal/Digital/Structure channel and 213 thousand trainable parameters in the Name channel. As for DWY100K datasets, there are about 2.6 million trainable parameters in the Literal/Digital/Structure channel and 213 thousand trainable parameters in the Name channel.
\subsubsection{Results Analysis}
The results of the proposed RAEA model and two groups of baseline models are displayed in Table \ref{Table_result}. The 4 models above the first solid line are Translation-based methods. The 8 models above the second solid line are the GNN-based methods. The number underlined is the maximum value in baselines. Below the second solid line, the results of RAEA are displayed in three versions with different ensemble strategies. The maximum value in each column is bolded. Based on the experimental results, we obtain the following insights:
\begin{enumerate}
\item[$\bullet$]
Among models used as baselines, GNN-based models generally have better results than Translation-based models, except for TransEdge. Initializing the entity embeddings randomly, TransEdge implements outstanding performance, which contextualizes the relation embeddings in terms of the specific entity pairs and interprets them as translations. We conclude that GNN-based models are more competitive in the entity alignment problem.
\item[$\bullet$]
Compared with baselines, the REAE implements a significant improvement on EA task, except for performance on the DWY100K datasets. As for the DBP15K dataset, the proposed RAEA model almost has an absolute advantage over baselines. Because the full utilization of alignment signals from attributes, relations, and interactions between them makes RAEA generate closer embedding for the same entities on a cross-lingual dataset, such as DBP15K. For the DWY100K, RAEA's performance is close to these recent advanced models which nearly achieve perfect results.
\item[$\bullet$]
The RAEA model with pre-weighted strategy implements promotion in all metrics and all datasets. This demonstrates that the ensemble strategy plays a significant role on the final results by controlling the weights of different channel, which carries different views of entity alignment signals. For the models that explore the EA task from separate perspectives of alignment signals on knowledge graphs, optimizing the ensemble strategy can effectively improve the performance of model on EA task.
\end{enumerate}
\subsubsection{Strengths of RAEA model}
To gain a deep insight into RAEA, we summarize the following three strengths of RAEA model.
\begin{enumerate}
\item[$\bullet$]\textbf{\itshape{Interactions between Attributes and Relations.}} The underlying idea of RAEA is to take the interactions of attribute triples and relation triples into consideration. We mainly achieve the idea with two crucial operations. Firstly, we generate the entity embedding by aggregating the attribute information. Secondly, we represent the relations with entity embeddings that have aggregated the attributes information and generate the entity embeddings again with relation representations. To have a deep insight into this idea, we compare our model with MultiKE\cite{zhang2019multi}. MultiKE generates entity embedding from different views of name, relations, and attributes respectively and combine them to get the ultimate entity embedding, where the attribute embeddings never join the calculation of the relation embeddings. However, in our RAEA model, we generate entity representation with attributes embedding and calculate the entity embedding and relation embedding iteratively, where the interactions between attribute triples and relation triples are involved. RAEA model obviously performs better than MultiKE, thus the usage of interactions between attributes and relations is one of the strengths of RAEA model.
\item[$\bullet$]\textbf{\itshape{The Diversity of Relation Types.}} Another advantage of RAEA model is that entity representation benefits from the diversity of relation types between two entities. To verify this point, we compare RAEA with AttrGNN\cite{liu2020exploring}. Although the attributes are aggregated in the entity embedding, the multi-relation between two entities is simplified as monotonous neighborship by generating the embedding of target entity with its neighbors. However, we generate a representation for each relation respectively in the relation-aware aggregator in RAEA model, reserving the diversity of relation types between two entities. Compared with the performance of AttrGNN, the proposed RAEA achieves remarkable improvement.
\item[$\bullet$]\textbf{\itshape{More Effective Ensemble Strategy.}} We conduct experiments on the RAEA model with the existing ensemble strategy, Average Pooling and SVM, and the new ensemble strategy we proposed, called Pre-weighted. Experimental results show that the Pre-weighted strategy implements 3.79\% improvement in DBP15K(zh-en), 5.5\% improvement in DBP15K(ja-en), 9.15\% improvement in DBP15K(fr-en), 2.38\% improvement in DWY100K(wd) on Hits@1, and 0.36\% improvement in DWY100K(yg) on Hits@1, compared with RAEA using Average Pooling or SVM ensemble strategy. This indicates that how to ensemble the similarity matrices that contain different alignment signals from different channels has important effects on EA task, especially in DBP15K.
\end{enumerate}

\begin{table*}[ht]
\centering
\setlength\tabcolsep{4pt}
\begin{tabular}{l|ccc|ccc|ccc|ccc|ccc} 
\toprule
\multirow{2}{*}{Methods}    & \multicolumn{3}{|c}{DBP15K (zh-en)} & \multicolumn{3}{|c}{DBP15K (ja-en)}  & \multicolumn{3}{|c}{DBP15K (fr-en)}& \multicolumn{3}{|c}{DWY100K (wd)}&\multicolumn{3}{|c}{DWY100K (yg)}\\& H@1         & H@10        & MRR            & H@1         & H@10        & MRR & H@1 & H@10        & MRR & H@1 & H@10        & MRR & H@1         & H@10        & MRR\\ 
\hline
w/o Attribute &78.02	&93.06	&0.835	&87.42	&96.92	&0.912	&94.37	&99.27	&0.960	&96.84	&99.42	&0.979	&99.55	&99.90	&0.997\\
w/o Relation &84.15	&95.41	&0.883	&87.68	&97.45	&0.913	&94.68	&99.44	&0.963	&97.31	&99.62	&0.981	&\textbf{99.97} &\textbf{100.00}	&\textbf{0.999}\\
w/o Name &72.17	&89.75	&0.787	&69.20	&88.03	&0.767	&58.84	&85.66	&0.683	&91.50	&97.95	&0.940	&99.31	&99.91	&0.996\\
w/o RGAT	&80.34	&94.17	&0.855	&87.15	&97.43	&0.911	&93.35	&99.34	&0.957	&96.86	&99.63	&0.980	&99.89	&\textbf{100.00}	&\textbf{0.999}\\
w/o MPnet	&83.28	&93.71	&0.872	&86.55	&96.35	&0.902	&94.14	&98.89	&0.960	&87.24	&94.53	&0.899	&98.77	&99.69	&0.991\\
\hline
RAEA	&\textbf{86.28}	&\textbf{97.07}	&\textbf{0.903}	&\textbf{88.48}	&\textbf{97.71}	&\textbf{0.920}	&\textbf{94.97}	&\textbf{99.60}	&\textbf{0.968}	&\textbf{97.35}	&\textbf{99.72}	&\textbf{0.983}	&99.82	&99.98	&\textbf{0.999}\\
\bottomrule
\end{tabular} 
\caption{The contrast of RAEA model with its variants on cross-lingual dataset DBP15K and monolingual dataset DWY100K, all experiments in the table use the proposed pre-weighted ensemble strategy.}
\label{contrast}
\end{table*}

\subsubsection{RAEA vs. Competitors}
To ensure fairness of comparison, we compare the performance of RAEA with GNN-based models that report experimental results on both DBP15K and DBP100K datasets in their original articles. In other words, we make analytical comparisons with GNN-based models in Table \ref{Table_result}. Comparisons between RAEA and MultiKE\cite{zhang2019multi}, AttrGNN\cite{liu2020exploring} have been discussed in the last section. AliNet\cite{sun2020knowledge} and Dual-AMN\cite{10.1145/3442381.3449897} enhance the similarity of two neighborhood structures by leveraging multi-hop neighbors, but the attributes are excluded in the neighbourhood and the diversity of relation type is ignored. Compared with them, RAEA enhances the implicit influence of two-hop neighbors with one layer of graph attention network. MuGNN\cite{Cao2019MultiChannelGN} distinguishes the multi-relations between two entities by generating a weight coefficient for each edge between to nodes, but the attributes are not involved in the weight calculation of edges. In comparison with MuGNN\cite{Cao2019MultiChannelGN}, RAEA integrates the entity embeddings into the representation for each relation to retain the relation richness between two entities. Different from RAEA, RDGCN\cite{Wu2019RelationAwareEA} only uses GCNs to incorporate alignment signals from neighboring structures, but not aggregate attributes information to entity embeddings. NMN\cite{Wu2020NeighborhoodMN} generates entity embeddings via neighborhood aggregation. Moreover, RKDEA\cite{RKDEA_2022} integrates neighborhood information and relational knowledge with knowledge distillation. However, both of them ignore the attributes when aggregating the neighborhood information. Leveraging the attribute, relations, and their interactions, RAEA implements better performance than the competitive GNN-based models on almost all evaluation metrics and all datasets.

\subsection{Ablation Study}
To analyze the effect of different components of RAEA model, we conduct an ablation study on the variants of RAEA as follows:
\begin{itemize}
\item[-] w/o Attribute: Ignore the attributes. The RAEA model without Literal/Digital channel.
\item[-] w/o Relation: Ignore the relations. The RAEA model without Structure channel.
\item[-] w/o Name: Ignore the entity names. The RAEA model without Name channel.
\item[-] w/o MPnet: The RAEA model without MPnet attribute encoder. We initialize the embeddings of literal/digital attributes and entity names with basic BERT.
\item[-] w/o RGAT: The RAEA model without Relation-aware Graph Attention Networks.
\end{itemize}

According to the Table \ref{contrast}, we find that there are three major observations:
\begin{enumerate}
\item[$\bullet$]The variant w/o Name performs obviously worse than the variant w/o Attribute and w/o Relation. This demonstrates the importance of entity names and the potential effect of involving entity names into EA task.
\item[$\bullet$]The variant w/o MPnet attribute encoder performs worse than RAEA with MPnet attribute encoder, which indicates that the embedding initialization of literal/digital attributes and entity names are important to EA task, though MPnet attribute encoder only is used just once before training each channel.
\item[$\bullet$]The full RAEA model outperforms the variant w/o Attribute and w/o Relation, which demonstrates the necessity of involving both attributes and relations in EA task.
\item[$\bullet$]RAEA performs better than its other variants, except for the DWY100K(yg) dataset, because the overall performance on DWY100K(yg) is approaching perfection. 
\end{enumerate}
From the ablation study, we conclude that attribute triples, relation triples and entity names are necessary to be considered for EA. And the Relation-aware graph attention networks and MPnet encoder contribute to the improvements in EA task. The good performance of RAEA model is largely attributed to Relation-aware graph attention networks, which have the capability of leveraging the interactions between attributes triples and relation triples by iteratively learning entity embeddings and relation embeddings.

\section{Conclusion}
We employ a two-stage pipeline to match identical or similar products that originate from two different e-commerce platforms, eBay and Amazon. In the Rough Filtering stage, the products are roughly filtrated based on rules defined by categories and titles of products. In the Fine Filtering stage, we propose RAEA model to find the Top-K matched Amazon products for each eBay product. 
The RAEA model adequately utilizes both attribute triples and relations triples so that the interactions between entities and relations are exactly captured for more reasonable entity representations. Different from AttrGNN\cite{liu2020exploring}, we use a new strategy (Pre-weighted) to ensemble different channels, which makes the channels complement each other better. Besides, we replace the original BERT encoder with MPnet attribute encoder to exact more informative features from text data. With the proposed RAEA model applied in the two-stage pipeline, we achieve impressive results on the eBay-Amazon dataset, where NDCG = 0.566 when we match Top-10 best matched Amazon products for each eBay product.
To prove the effectiveness of RAEA model, we compare its performance with baseline models on public datasets DBP15K and DWY100K for EA task, which achieves state-of-the-art performance.\\

\noindent \textbf{Ethical Approval }\\
No ethical issue involved.\\

\noindent \textbf{Competing interests }\\
No competing interest.\\

\noindent \textbf{Authors' contributions}\\
QZ and HY initialized the project and managed the study. WL, JP, XZhang, XG, YY and XZhao collected the data and performed formal analysis. All authors analyzed the data. WL and QZ wrote the initial manuscript. All authors contributed to the editing of the paper.\\

\noindent \textbf{Funding }\\
This study was supported partly by the National Key Research and Development Program of China (No. 2019YFE0198600), and partly by InnoHK initiative, the government of the HKSAR, and the Laboratory for AI-Powered Financial Technologies.\\

\noindent \textbf{Availability of data and materials }\\
All datasets are open-source and the sources are cited.

\bibliographystyle{ijcaArticle.bib}
\bibliography{BibTex-Sample}

\end{document}